\documentclass[runningheads]{llncs}
\usepackage{graphicx}
\usepackage{tabularx}
\usepackage{comment}
\usepackage{amsmath} 
\usepackage{color}
\usepackage{graphicx}
\usepackage{amssymb}
\usepackage{booktabs}
\usepackage{hhline}
\usepackage{multirow}
\usepackage{multicol}
\usepackage{microtype}      
\usepackage{xcolor}         
\usepackage{hhline}
\usepackage{bbm}
\usepackage{boldline}
\usepackage{amssymb}
\usepackage{algorithm}
\usepackage{algpseudocode}
\usepackage{hyperref}
\hypersetup{
    colorlinks=true,
    urlcolor=blue,
}
\usepackage{pifont}
\usepackage{soul} 
\newcommand{\cmark}{\ding{51}}%
\usepackage{wrapfig}
\usepackage{tikz}
\usepackage{cite}
 
\usepackage[accsupp]{axessibility}  

\DeclareMathOperator*{\argmax}{arg\,max}

\begin{document}
\pagestyle{headings}
\mainmatter
\def\ECCVSubNumber{6283}  

\title{Difficulty-Aware Simulator \\ for Open Set Recognition} 

\titlerunning{Difficulty-Aware Simulator for Open Set Recognition}
%
\author{WonJun Moon \and Junho Park \and Hyun Seok Seong \and Cheol-Ho Cho \and Jae-Pil Heo\thanks{Corresponding author}\\}
\authorrunning{WJ. Moon et al.}
%
\institute{Sungkyunkwan University\\
\email{\{wjun0830,pjh4993,gustjrdl95,gersys,jaepilheo\}@skku.edu}}
\maketitle
\begin{abstract}
Open set recognition (OSR) assumes unknown instances appear out of the blue at the inference time.
The main challenge of OSR is that the response of models for unknowns is totally unpredictable.
Furthermore, the diversity of open set makes it harder since instances have different difficulty levels.
Therefore, we present a novel framework, DIfficulty-Aware Simulator (DIAS), that generates fakes with diverse difficulty levels to simulate the real world.
We first investigate fakes from generative adversarial network (GAN) in the classifier's viewpoint and observe that these are not severely challenging.
This leads us to define the criteria for difficulty by regarding samples generated with GANs having moderate-difficulty.
To produce hard-difficulty examples, we introduce Copycat, imitating the behavior of the classifier. 
Furthermore, moderate- and easy-difficulty samples are also yielded by our modified GAN and Copycat, respectively.
As a result, DIAS outperforms state-of-the-art methods with both metrics of AUROC and F-score.
Our code is available at \href{https://github.com/wjun0830/Difficulty-Aware-Simulator}{https://github.com/wjun0830/Difficulty-Aware-Simulator}.
\end{abstract}

\section{Introduction}
\label{sec:intro}
Thanks to the advance of convolutional neural network (CNN), downstream tasks of computer vision have been through several breakthroughs~\cite{he2016deep, krizhevsky2012imagenet}. 
Although the performance of deep learning is now comparable to that of humans, distilling knowledge learned from known classes to detect unseen categories lags behind~\cite{guo2021caps, girish2021towards}. 
Hence, unseen categories are often misclassified into one of the known classes.
In this context, open set recognition was proposed to learn for the capability of detecting unknowns~\cite{bendale2015towards, bendale2016openmax, scheirer2012toward}.

\begin{figure}
    \centering
    \includegraphics[width=0.55\textwidth]{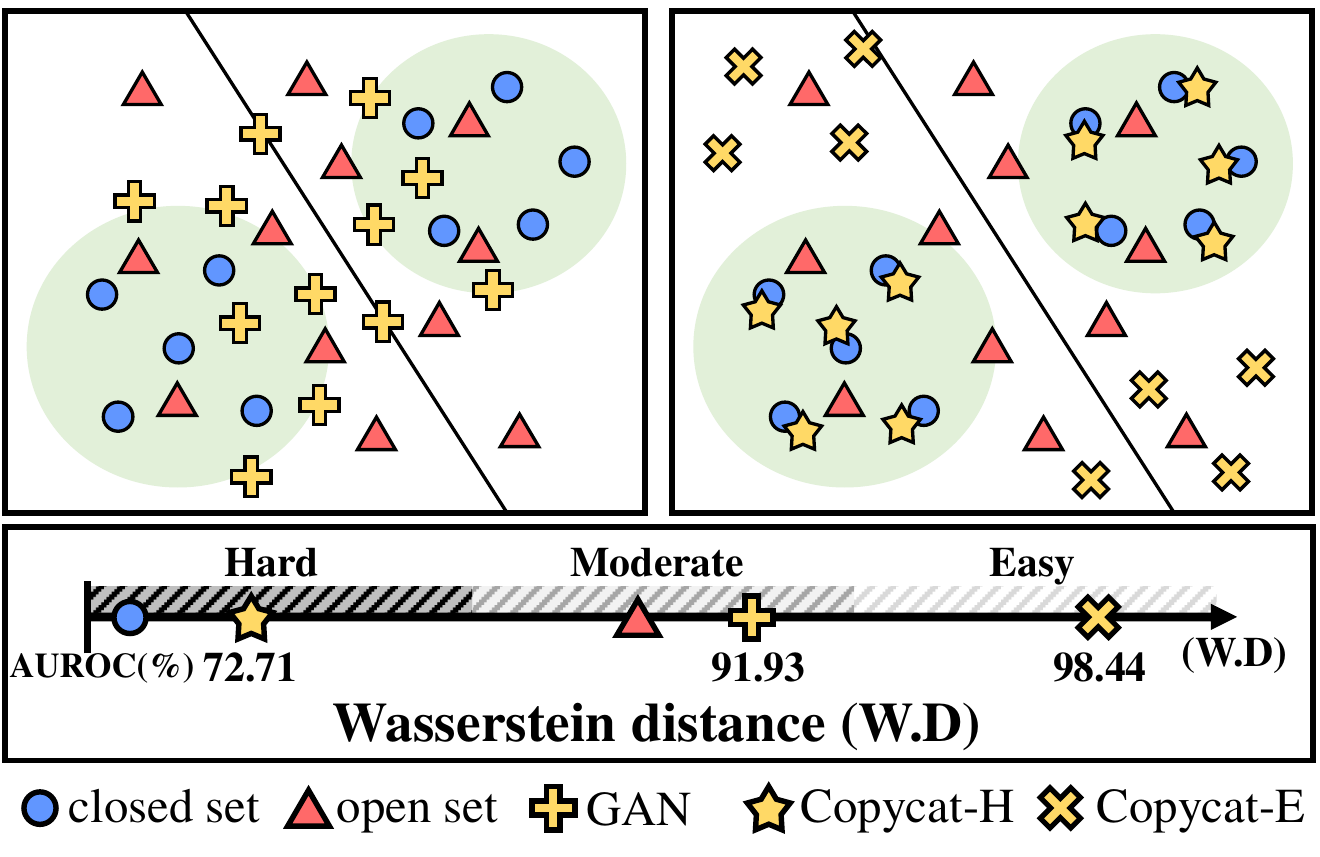}
    \caption{Given a set of closed and open classes, we intuitively describe how the classifier recognizes fake examples from the image-generator and the Copycat.
    Green circles indicate the confident class boundaries.
    (Left) Both open set and generated instances by GAN usually project nearby the class boundaries or sometimes inside.
    (Right) Hard-fake instances produced by the Copycat embed deep inside the class boundaries, while easy ones usually embed far.
    (Bottom) Each set of fakes are represented on a line based on normalized Wasserstein distances (W.D) to closed set with the corresponding AUROC score. To measure W.D and AUROC scores, we conduct a primary experiment on CIFAR10 with the classifier trained only on closed set.
    In this paper, we define the difficulty levels according to two measures from the perspective of the classifier.
    }
    \label{figure_motivation}
\end{figure}
Generally, CNN are often highly biased to yield high confidence scores~\cite{perera2020gdfr}.
However, calibrating the confidence to distinguish open set data is infeasible due to the inaccessibility of those unseen data in the training phase~\cite{guo2017calibration}.
Furthermore, it is known that the learned CNN classifiers tend to highly rely on the discriminative features to distinguish classes but ignore the other details~\cite{roady2020difsim}.
Therefore, open set instances that share such discriminative features with closed set can be easily confused.
In other words, open set may have a level of difficulty, which is determined according to the degrees of feature sharing with the closed set. 
In this regard, it is challenging to cope with all open set with various difficulty levels which can be encountered in the real world.
Due to the inaccessibility to open set during training, substantial efforts were put to simulate virtual open set~\cite{ge2017gopenmax, neal2018osrci, chen2020rpl, chen2021arpl, zhou2021proser}. 
RPL~\cite{chen2020rpl} applied 1-vs-rest training scheme and exploited features from other classes to form open set space, and
PROSER~\cite{zhou2021proser} mixed features to simulate open set. 
Furthermore, GAN~\cite{goodfellow2014gan} is actively used to synthesize unknown samples~\cite{ge2017gopenmax, neal2018osrci, chen2021arpl}.
However, diverse difficulty levels of open set are not taken into account.
Specifically, feature simulation methods only utilize the features outside the class boundaries which are easy to be distinguished~\cite{chen2020rpl, zhou2021proser}.
Besides, image generation-based methods mostly produce samples being predicted as unknown class by the classifier to represent open set~\cite{ge2017gopenmax, lee2017training, neal2018osrci, chen2021arpl}.
These samples hardly have high difficulty, so the classifiers learned with them can be still vulnerable to difficult open set that contains semantically meaningful features of one of the known classes~\cite{nguyen2015deep, geirhos2020shortcut}.
In this context, we propose a novel framework, DIfficulty-Aware Simulator (DIAS), that exposes diverse samples to the classifier with various difficulty levels from the classifier's perspective.
As shown in Fig.~\ref{figure_motivation}, we found that a set of generated images with GAN is not very challenging for the classifier.
Therefore, with the GAN as a criterion, we define the difficulty levels and introduce the Copycat, a feature generator producing hard fake instances. 
As the training iteration proceeds, the Copycat mimics the behavior of the classifier and generates real-like fake features that the classifier will likely yield a high probability.
In other words, the classifier faces unknown features within its decision boundaries at every iteration.
In this way, the classifier is repeatedly exposed to confusing instances and learns to calibrate even within the class boundaries.
Moreover, we further ask the Copycat to create easy fake samples and also modify the image-level generator to take the classifier's perspective into account.
These fake instances are additionally utilized to simulate the real world in which unseen examples with various difficulties may exist.
Besides, DIAS is inherently equipped with a decent threshold to distinguish open set.
It enables to avoid expensive process to search an appropriate confidence threshold of the classifier for OSR.

In summary, our contributions are:
(i) We propose a novel framework, DIAS, for difficulty-aware open set simulation from the classifier's perspective. To the best of our knowledge, this is the first attempt to consider the difficulty level in OSR.
(ii) We present Copycat, the difficult fake feature generator, by imitating the classifier's behavior with the distilled knowledge.
(iii) We prove effectiveness with competitive results and demonstrate feasibility with an inherent threshold to identify open set samples.

\section{Background and Related Works}
\label{sec:related}
\textbf{Open set recognition}
To apply the classification models to real world with high robustness, OSR was first formalized as a constrained minimization problem~\cite{scheirer2012toward}.
Following them, earlier works used traditional approaches: support vector machines, Extreme Value Theory (EVT), nearest class mean classifier, and nearest neightbor~\cite{jain2014multi, scheirer2014probability, bendale2015towards, zhang2016sparse, junior2017nearest}.
Then, along with the development in CNN, deep learning algorithms have been widely adopted.
In the beginning, softmax was tackled for its closed nature.
To replace this, Openmax~\cite{bendale2016openmax} tried to extend the classifier and K-sigmoid~\cite{shu2017doc} conducted score analysis to search for threshold.

A recently popular stream for OSR is employing generative models to learn a representation that only preserves known samples.
Conditional Variational AutoEncoder (VAE) was utilized in C2AE~\cite{oza2019c2ae} to model the reconstruction error based on the EVT.
CGDL~\cite{sun2020cgdl} improved the VAE's weakness in closed set classification with conditional gaussian distribution learning.
Moreover, flow-based model was employed for density estimation~\cite{zhang2020hybrid} and capsule network was adopted to support representation learning with conditional VAE~\cite{guo2021caps}.
Other approaches exploited the generative model's representation as an additional feature.
GFROSR\cite{perera2020gdfr} employed reconstructed image from an autoencoder to augment the image while CROSR\cite{yoshihashi2019crosr} adopted ladder network to utilize both the prediction and the latent features for unknown detection.

Other methods mostly fall into the category of simulating unknown examples, a more intuitive way for OSR.
RPL~\cite{chen2020rpl} tried to conduct simulation with prototype learning. 
With prototypes, they designed an embedding space for open set at the center where the samples will yield low confidence scores.
Then, based on manifold mixup~\cite{verma2019manifold}, PROSER~\cite{zhou2021proser} set up the open space between class boundaries to keep each boundary far from others.
GAN was also employed to simulate open set. 
G-openmax~\cite{ge2017gopenmax} improved openmax~\cite{bendale2016openmax} via generating extra images to represent the unknown class and OSRCI~\cite{neal2018osrci} developed encoder-decoder GAN architecture to generate counterfactual examples.
Additionally, ARPL~\cite{chen2021arpl} enhanced prototype learning with generated fake samples and GAN was further extended to feature space~\cite{kong2021opengan}.
DIAS shares similarity with these methods in that we simulate open set.
However, the main difference comes from the consideration of difficulty gaps between open set instances from the classifier's perspective.

\textbf{Multi-level knowledge distillation}
Knowledge Distillation (KD) was introduced in \cite{hinton2015distilling} where the student learns from the ground-truth labels and the soft-labels from the teacher.
AT~\cite{zagoruyko2016paying} exploited attention maps in every layer to transfer the knowledge and FSP~\cite{yim2017gift} utilized the FSP matrix that contains the distilled knowledge from the teacher. Moreover, cross-stage connection paths were also introduced to overcome information mismatch arising from differences in model size~\cite{chen2021distilling}.
Copycat share similar concept with multi-stage KD methods that it imitates encoding behavior of the classifier.
Up to we know, developing a fake generator with KD is a novel strategy in the literature of OSR.

\section{Methodology}
\subsection{Problem Formulation and Motivation}
\label{sec:problemformulation}


The configuration of OSR is different from classification since models can face unseen objects at the inference time.
Suppose that a model trained with $\mathcal{D}_{tr} = \left ( X, Y \right )$ over a set of classes $K$. $X$ is a set of input data $\mathbf{x}$ and $Y$ is a set of one-hot labels which each sample $\mathbf{y} \in {\{0,1\}}^{K}$, where its value is $1$ for the ground-truth class and 0 for the others. 
A typical classification evaluates the trained model on $\mathcal{D}_{te}=\left ( T, Y \right )$ where $X$ and $T$ are sampled from the same set of classes.

On the contrary, $\mathcal{\hat{D}}_{te} = ( \hat{T} )$ of OSR contains instances over a novel category set $\hat{K}$. 
Since the conventional classifier assumes $K$ categories, common approach is to distinguish open set based on confidence thresholding~\cite{hendrycks2016baseline}. 
However, challenges in OSR come from the similarity between $\mathcal{D}_{te}$ and $\mathcal{\hat{D}}_{te}$ which often leads to high confidence scores for both. 
Furthermore, the diverse relationships between $\mathcal{D}_{te}$ and $\mathcal{\hat{D}}_{te}$ affect the threshold to be vulnerable to different datasets.
To this end, the classifier is preferred to be calibrated to yield low score on unknown classes.

Intuitive approach to calibrate the classifier is to simulate the $\mathcal{\hat{D}}_{te}$ with fake set $\bar{X}$. 
Then, while the classifier is trained on $\mathcal{D}_{tr}$, it is also enforced to suppress the output logit of the fake sample $\mathbf{\bar{x}}$ to a uniform probability distribution:
\begin{equation}
\label{eq:ce+sce}
    \mathcal{L} = \mathcal{L}_{ce}(\mathbf{x}, \mathbf{y}) + \lambda \cdot \mathcal{L}_{ce}(\mathbf{\bar{x}}, 1/K\cdot\mathbf{u}),
\end{equation}
where $\mathcal{L}_{ce}$, $\lambda$, and $\mathbf{u}$ each denotes cross-entropy loss, scale parameter for fake sample calibration, and the all-one vector.


Nevertheless, due to the nature of the real world where unknown classes are rampant, a single set of fakes cannot represent all unknowns~\cite{godin}.
This is because the difficulty levels of data in the perspective of the classifier can be significantly varying~\cite{tudor2016hard, russakovsky2015imagenet}.
This motivates us to develop a simulator to rehearse with fake examples in diverse difficulty levels.
To overcome the new challenge in generating the hard-fake instances, we first introduce Copycat learning which generates hard fakes in Sec.~\ref{sec:copycat} and describe the DIAS in Sec.~\ref{sec:DIAS}.
Finally, we explain the inherent threshold of DIAS which benefits inference procedure in Sec.~\ref{sec:inference}
\begin{figure*}[t]
    \centering
    \includegraphics[scale=0.42]{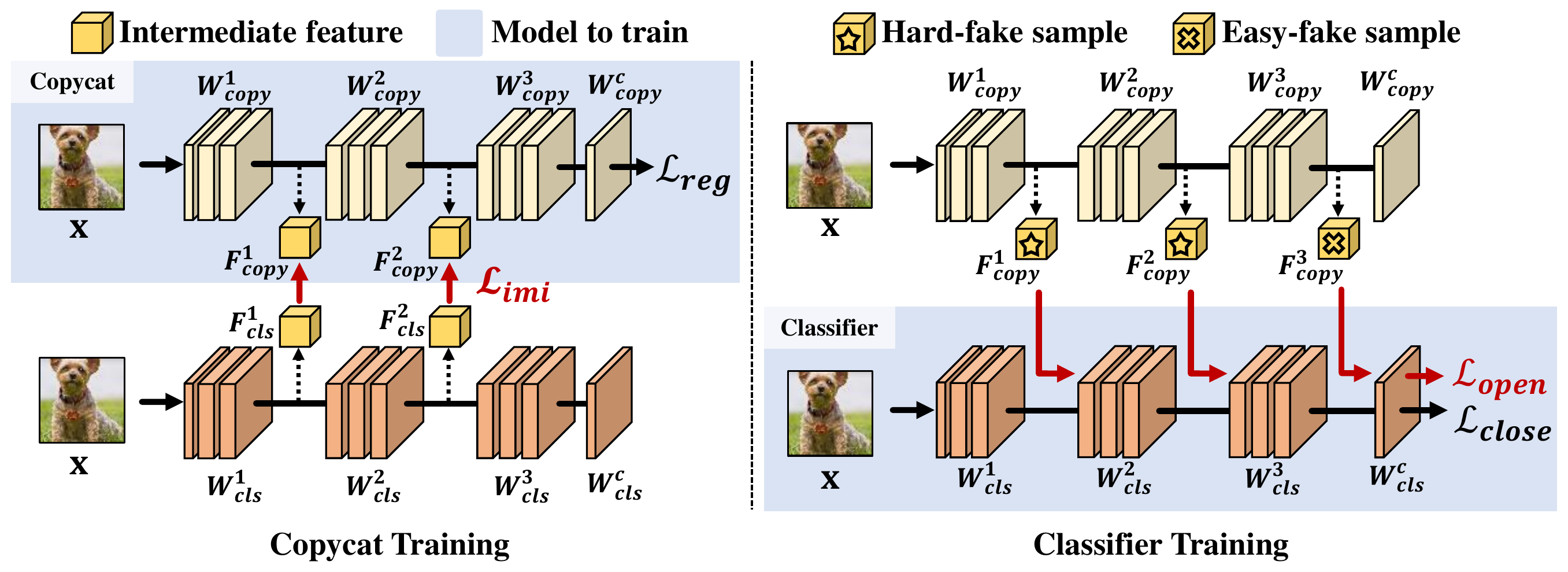}
    \caption{Illustration of the joint training scheme between the Copycat and the classifier. (Left) Knowledge is distilled from the classifier to the Copycat while its parameters are also updated by the regularization loss. (Right) The classifier is exposed to fakes generated by the Copycat. Depending on the existence of the imitation loss $\mathcal{L}_{imi}$ in its group ($W^{1}, W^{2}, W^{3}$), each convolutional group of the Copycat outputs hard- or easy- fake instances. 
    $\mathcal{L}_{open}$ is a general term for the loss computed from fake examples.
    }
    \label{figure_framework}
\end{figure*}
\subsection{Copycat : Hard-fake feature generator}
\label{sec:copycat}
In Fig.~\ref{figure_framework}, we illustrate how the Copycat interacts with the classifier to introduce fake examples.
Note that, throughout the paper, the terms regarding the difficulty level (e.g. Hard, Moderate, and Easy) are the data difficulty from the perspective of the classifier, as in Fig.~\ref{figure_motivation}.
Given an input image $\mathbf{x}$ and network $W$, we let $\mathbf{p} = W(\mathbf{x})$ stand for the output probability. $W$ can be separted into different parts $(W^{1},\cdots ,W^{n} ,W^{c})$, where $W^{c}$ is the fully-connected layer with the softmax activation and $W^{1},\cdots,W^{n}$ are different groups of layers separated by predefined criteria. Then, predicting $\mathbf{p}$ with network $W$ can be expressed as
\begin{equation}
\label{eq_general_network}
    \mathbf{p} = W^{c} \circ W^{n} \circ \cdots \circ W^{1}(\mathbf{x}).
\end{equation}
We refer to ``$\circ$" as nesting of functions where $g \circ f(x) = g(f(x))$ .
Note that the overall architecture for the Copycat and the classifier is equivalent and layers are grouped with the same criteria.
For simplicity, we split the Copycat and the classifier into three groups of layers.
This can be easily applied to other models~\cite{he2016resnet, huang2017densenet} as the number of groups can be adjusted.
We also let intermediate features to be denoted as $(F^{1}, \cdots, F^{n})$. For instance, $i$th feature is calculated as:
\begin{equation}
\label{eq_per_level_feature}
    F^{i} = W^{i} \circ W^{i-1} \circ \cdots \circ W^{1}(\mathbf{x}).
\end{equation}





The key objective of the Copycat is to create virtual features of hard- and easy-difficulty levels.
To introduce the training procedure, we first define $I$ as an index set for convolutional groups which is subdivided into $I_{hard}$ and $I_{easy}$.
$I_{hard}$ and $I_{easy}$ imply the difficulty level of fake features that each convolutional group outputs.
Then, we train the Copycat to mimic the classifier's knowledge at $I_{hard}$ and to differ from the classifier at $I_{easy}$ with the loss function formulated as:
\begin{equation}
\label{eq_Copycat}
    \mathcal{L}_{copy} = \mathcal{L}_{reg} + \mathcal{L}_{imi},
\end{equation}
where $\mathcal{L}_{reg}$ denote a regularization loss and $\mathcal{L}_{imi}$ is an imitation loss.
Note that while $\mathcal{L}_{reg}$ is for all layers in the Copycat, $\mathcal{L}_{imi}$ is only updated to layers in $I_{hard}$.
Therefore, the Copycat gets to behave similar to the classifier at $I_{hard}$.

Without loss of generality, we state that the imitation losses are placed between the features of the Copycat and the classifier where they are of the same convolutional group index.
We define the imitation loss as:
\begin{equation}
\label{eq_imitation}
    \mathcal{L}_{imi} = \sum_{j\in I_{hard}}\|F_{copy}^{j} - F_{cls}^{j}  \|_{1},
\end{equation}
where $F_{copy}^{j}$ and $F_{cls}^{j}$ are the feature vectors from $j$-th convolutional group of the Copycat $W_{copy}^{j}$ and the classifier $W_{cls}^{j}$.
Then, with $\mathcal{L}_{imi}$, forcing the convolutional groups at $I_{hard}$ of the Copycat to behave similarly to classifier's, hard layers in the Copycat become to produce difficult fake features from the classifier's perspective.
However, if $I_{easy}$ is defined in front of $I_{hard}$, corrupted features from $I_{easy}$ can lead to unstable $\mathcal{L}_{imi}$ because $I_{easy}$ is to yield abnormal features that are different from the ones of the classifier.
Thus, as the quality of hard fakes is crucial in the Copycat, we define $I_{easy}$ at the last. 


Regularization loss has two purposes: hindering the replication procedure to prevent Copycat from being exactly the same as the classifier and diversifying easy fakes. 
Since the groups at $I_{easy}$ of the Copycat do not have any connectivity to the classifier and are updated only with $\mathcal{L}_{reg}$, features from $I_{easy}$ of the Copycat would be abnormal which are easy to be identified by the classifier.
Moreover, as $\mathcal{L}_{reg}$ is iteratively applied, diverse easy fakes are produced.
For $\mathcal{L}_{reg}$, we simply use $\mathcal{L}_{ce}$ with real labels since the classification loss plays two roles well. 





\subsection{Difficulty-Aware Simulator}
\label{sec:DIAS}
The key idea of DIAS is to expose the classifier with open set of various difficulty levels.
To consider the classifier's perspective in real-time, we apply joint scheme of the training phase between each generative model and the classifier.

\noindent
\textbf{Stage \textbf{I} : Generator Training.}
To come up with virtual open set for simulation, we employ a Copycat and an image-generator.
As we discussed in Sec.~\ref{sec:copycat}, the Copycat is to prepare hard- and easy-fake features for robust training of the classifier.
Moreover, inspired by Fig.~\ref{figure_motivation}, we employ GAN to generate moderate-level fake images but with a little modification to consider the classifier's viewpoint.
\begin{figure}[t]
    \centering
    \includegraphics[scale=0.43]{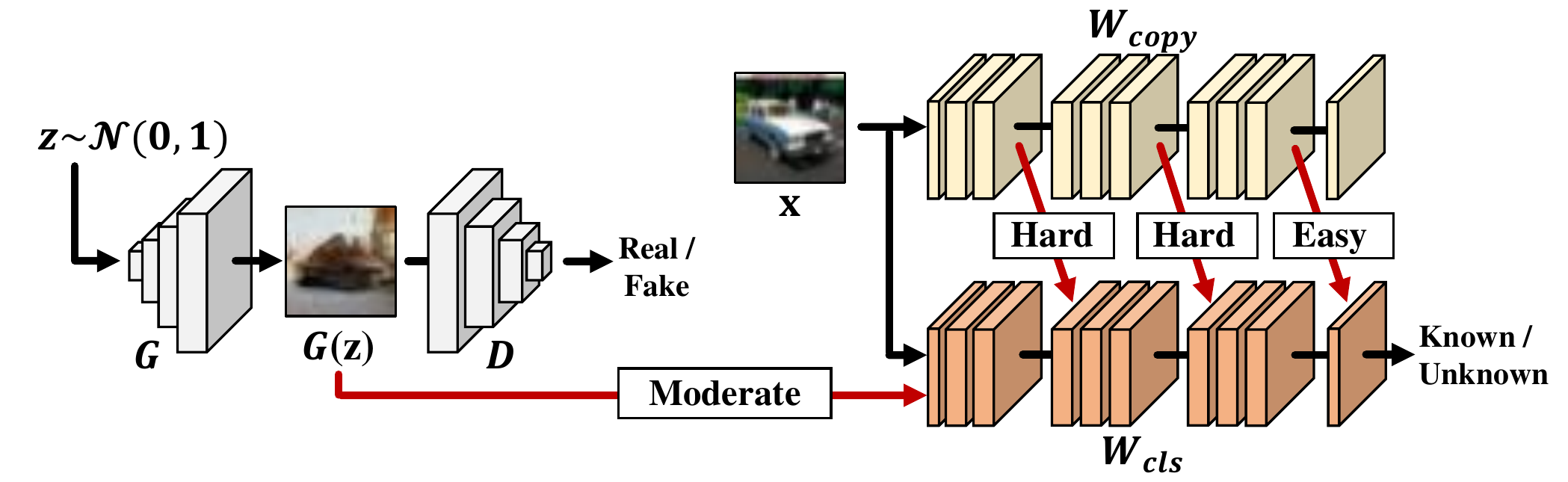}
    \caption{Illustration of the proposed DIAS. Our GAN and Copycat each receives the noise vector $z$ and input image $x$ to produce fake instances with various difficulties. These instances are provided to the classifier to have a rehearsal for OSR.
    }
    \label{fig:DIAS}
\end{figure}

To synthesize fake images of the moderate difficulty, the generator is adversarially trained with the discriminator. 
Let $D$ and $G$ be the discriminator and the generator, respectively.
The generator receives the noise vector $\mathbf{z}$ sampled from normal distribution $\mathcal{N}$ and generates the output as $G(\mathbf{z})$.
Then, the discriminator encodes both the image $\mathbf{x}$ and fake images $G(\mathbf{z})$ to the probability $[0, 1]$ whether to predict the input is from a real distribution or not.

Furthermore, our focus of generating fake images is to prepare the classifier for open set.
Thus, optimization process for the generator should consider the classifier's prediction.
However, arranging the process should be carefully designed to avoid two situations.
On the one hand, when the perfect generator is trained, the classifier would not be capable of discriminating open set due to the equilibrium in the mini-max game~\cite{kong2021opengan, arora2017generalization}.
On the other hand, when the generator only outputs images that the classifier predicts with the uniform distribution, generated samples would not be representative for simulating moderate-difficulty open set.
Therefore, we encourage the generator to output closed set-like images, but very cautiously, with the classifier's predictions.
Specifically, we use the negative cross-entropy loss to a uniform target vector.
Accordingly, the generator is trained to produce fake images that are neither the known nor the outlier from the classifier’s perspective.
Formally, the overall optimization process of our GAN is conducted by:
\begin{equation}
\label{eq:gan}
    \begin{split}
    &\max_{D}\min_{G} \;\; \mathcal{L}_{gan} = \;{\mathbb E}_{\mathbf{x}\sim {\mathcal D}_{tr}} \big[ \log D(\mathbf{x}) \big]\\[-5pt]
    &+\mathbb{E}_{\mathbf{z} \sim  {\mathcal N}} \big[ \log \big(1- D(G(\mathbf{z})) \big) -\beta/K\sum^{K}_{k=1}\log C_{k} \big(G(\mathbf{z})\big) \big] \\[-5pt]
    \end{split}
\end{equation}
where $\beta$ is the scale parameter for negative cross-entropy and $C_{k}(\cdot)$ outputs probability of class $k$ for given input. 



\noindent
\textbf{Stage \textbf{II} : Classifier Training.}
Facing all kinds of difficulty through virtual open set at training time, we pursue to drive the classifier to be prepared for handling unseen classes.
Thus, we formulate the loss for the classifier as:
\begin{equation}
\label{eq:classifier}
    \mathcal{L}_{cls} = \mathcal{L}_{close} + \lambda \cdot \mathcal{L}_{open},
\end{equation}
where $\mathcal{L}_{close}$ and $\mathcal{L}_{open}$ are to discriminate within known and between known and unknown classes, respectively.
To consider the difficulty levels in the classifier's behavior, we need to differentiate the objectives with their difficulties.
Therefore, we employ the $\mathcal{L}_{ce}$ for both but with the smoothed label for calculating the $\mathcal{L}_{open}$.
The smoothed label $\mathbf{\Tilde{y}}$ is formed with smoothing ratio $\alpha$ as below:
\begin{equation}
\label{eq_labelsmoothing}
    \mathbf{\Tilde{y}} = (1 - \alpha)\cdot\mathbf{y} + \alpha/K\cdot\mathbf{u}.
\end{equation}

\begin{figure}[t]
    \centering
    \includegraphics[scale=0.35]{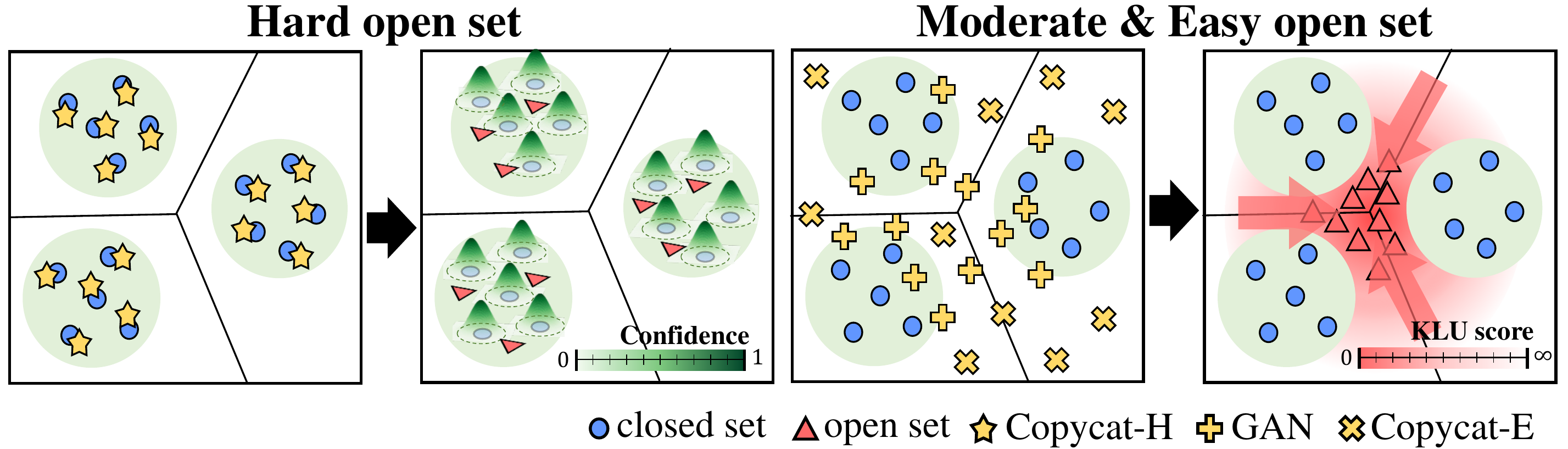}
    \caption{
    Illustration about the changes in the classifier's perspective by facing fake open set.
    Confidence bar represents $\mathbf{p}$ and KLU score bar stands for the KL-divergence to the uniform distribution.
    (Left) Hard-difficulty open set samples are mostly located within the decision boundary of the classifier. They lead the classifier to calibrate its find-grained confidence distribution to distinguish such hard-difficulty samples.
    (Right) Moderate- and easy-difficulty samples are likely located near the boundary or free space. The classifier tries to gather them to the region with the uniform probability.
    }
    \label{fig:behavior}
\end{figure}

Generated fakes are encoded by the classifier to make a prediction as illustrated in Fig.~\ref{fig:DIAS}.
Specifically, intermediate features from the Copycat and fake images from GAN are passed on to the classifier to predict $\mathbf{p}_{copy}^{i}$ and $\mathbf{p}_{gan}$ as: 
\begin{align}
    & \mathbf{p}_{copy}^{i} = W_{cls}^{c} \circ W_{cls}^{n} \circ \cdots \circ W_{cls}^{i+1}(F^{i}_{copy}) \label{eq:open_p}\\[-2pt]
    & \mathbf{p}_{gan} = W_{cls}^{c} \circ W_{cls}^{n} \circ \cdots \circ W_{cls}^{1}(G(z)) \label{eq:open_g}
\end{align}
Then, open loss is calculated alternately depending on the training phase:
\begin{equation}
    \label{eq:openloss}
    \begin{aligned}
        \mathcal{L}_{open} =
        \begin{cases}
        \sum_{i\in I} \mathcal{L}_{ce}(\mathbf{\Tilde{y}}, \mathbf{p}_{copy}^{i}) \\
        \sum_{z \sim \mathcal{N}} \mathcal{L}_{ce}(\frac{1}{K}\cdot\mathbf{u}, \mathbf{p}_{gan}) 
        \end{cases}
    \end{aligned}
\end{equation}
\begin{wrapfigure}{t!}{6.3cm}
\vspace{-1.5cm}
\begin{minipage}{0.52\textwidth}
\begin{algorithm}[H]
    \caption{Training DIAS}
    \label{algorithm_DIAS}
    \begin{algorithmic}[H]
        \Require Parameters of Classifier, Copycat, Generator and Discriminator $(\theta, \phi, \psi, \omega)$, Learning rate $\eta$
	    \For{$i \in \{1,..., epoch\}$}
	        \State {\color{gray}{$\triangleright$ Phase \textbf{I}: Training with Copycat}}
		    \State $\mathcal{L}_{copy} = \mathcal{L}_{imi} + \mathcal{L}_{reg}$
            \State $\phi \leftarrow \phi - \eta \nabla_{\phi}\mathcal{L}_{copy}$
		    \State $\mathcal{L}_{cls} = \mathcal{L}_{close} + \lambda \cdot \mathcal{L}_{open}$
            \State $\theta \leftarrow \theta - \eta \nabla_{\theta}\mathcal{L}_{cls}$
            \State {\color{gray}{$\triangleright$ Phase \textbf{II}: Training with GAN}}
            \State $\psi \leftarrow \psi - \eta \nabla_{\psi}\mathcal{L}_{gan}$
            \State $\omega \leftarrow \omega - \eta \nabla_{\omega}\mathcal{L}_{gan}$
		    \State $\mathcal{L}_{cls} = \mathcal{L}_{close} + \lambda \cdot \mathcal{L}_{open}$
            \State $\theta \leftarrow \theta - \eta \nabla_{\theta}\mathcal{L}_{cls}$
	    \EndFor    
    \end{algorithmic}
\end{algorithm}
\end{minipage}%
\vspace{-0.7cm}
\end{wrapfigure}
To define $\mathbf{\Tilde{y}}$ for each difficulty group, we grant a smaller value to $\alpha$ when the input has a high probability of belonging to one of the known categories.
To be specific, we divide $\alpha$ into $\alpha_{hard}$ and $\alpha_{easy}$. 
Note that motivated by ARPL~\cite{chen2021arpl}, we set the target label for the fake image of the GAN to a uniform distribution since it works better to regard them as unknown.
Likewise, we also set $\alpha_{easy}$ to 1 to further regard easy fakes as unknown.
On the other hand, since identifying hard fake instances as open set might contradict the training procedure, $\alpha_{hard}$ is set to 0.5.
Therefore, the easy and moderate fake examples are forced to reside outside the class boundaries, while a difficult set of fakes are implemented to be calibrated within the class boundaries.

In Fig.~\ref{fig:behavior}, we describe how diverse levels of difficulty assist the classifier to build robust decision boundaries.
As can be seen, while easy and moderate fake sets play a role of gathering open set to the parts where they can retain uniform distribution, hard fakes force the classifier to calibrate its decision boundaries with the smoothed label. Our training procedure is outlined in Algorithm~\ref{algorithm_DIAS}.

\subsection{Inherent threshold of DIAS}
\label{sec:inference}

DIAS has its benefits at the inference for threshold selection.
In general, cross-validation is a widely adopted strategy to specify threshold when identifying the open set, although it is time-consuming work to find an appropriate threshold.
On top of this, the value of threshold is very sensitive that it must be explored for every pair of known and unknown dataset~\cite{neal2018osrci}.
In contrast, our proposed DIAS is inherently equipped with a criterion to detect unknowns.
To be precise, we observe that $\max\limits_{k} \mathbf{\Tilde{y}}$ computed with $\alpha_{hard}$ works out to be the decent threshold.
This is because we enforce the confidence scores for virtual open set to be equal or lower than the target confidence for difficult fake samples $\max\limits_{k} \mathbf{\Tilde{y}}$.
Formally, given an image $\mathbf{x}$, we extend the closed set classifier by predicting label $\hat{y}$ as:
\begin{equation}
    \small
    \begin{aligned}
        \hat{y} =
        \begin{cases}
        K + 1 & \: \text{if} \: \max \big\{W (\mathbf{x})\big\} < \tau + \epsilon \\
        \argmax\limits_{k=1,\cdots,K} \: \big\{W (\mathbf{x}) \big\} & \: \text{otherwise}
        \end{cases}
    \end{aligned}
    \label{eq:inference}
\end{equation}
where $K + 1$ is a class for unknown and $\tau=(1 - \alpha_{hard}) + \alpha_{hard}/K$. 
Note that $\tau$ is same as simply calculating the maximum value in Eq.~\ref{eq_labelsmoothing} with $\alpha_{hard}$.
Therefore, we do not need any extra algorithm to compute the threshold.

\section{Experiments}
\label{sec:exp}

\subsection{Evaluation protocols, metrics, and datasets}
\textbf{Evaluation protocols.} With a $c$-class dataset, OSR scenario is generally designed by randomly selecting $K$ classes as known where $(c \gg K)$. 
Then, the remaining $c - K$ classes are considered as open set classes. 
Then, five-randomized scenarios are simulated to measure the Area Under the Receiver Operating Characteristic (AUROC) curve or F-score.
Note that the split information can be different over methods since they conduct experiments on randomized trials~\cite{neal2018osrci, oza2019c2ae, chen2020rpl, yue2021zsrosr}.
However, as different split often leads to unfair comparison, recent methods pre-define split information on each AUROC~\cite{guo2021caps, neal2018osrci, yoshihashi2019crosr} and F1-score benchmark~\cite{yue2021zsrosr}.
Following them, we conduct experiments on two standardized split information for a fair comparison. Split information is in the supplementary.

\noindent
\textbf{Metrics.} We use two metrics: F1-score and AUROC score. F1 is a more practical measure, representing the classification accuracy. It is calculated as a weighted average of precision and recall.
For OSR, F1-score is commonly obtained by adding an extra class for open set and searching for proper threshold.
On the other hand, AUROC is a metric that does not require any calibration process. It considers the trade-off between true positive rate and false positive rate across different decision thresholds.

\noindent
\textbf{Datasets.} Prior to dataset explanation, we describe the term, openness, which indicates the ratio between the known and the unknown:
\begin{equation}
\label{eq_openness}
    Openness = 1- \sqrt{K / (K + \hat{K})}
\end{equation}
where $K$ and $\hat{K}$ stand for the number of classes for known and unknown, respectively.
With openness, we discuss several benchmarking datasets:
\textbf{MNIST}, \textbf{SVHN}, \textbf{CIFAR10}~\cite{lecun1998mnist, netzer2011reading, krizhevsky2009learning} 
contain 10 classes. Six classes are chosen to be known and four classes are to be unknown classes. Openness is 22.54\%.
\textbf{CIFAR+10}, \textbf{CIFAR+50} are artificially synthesized with four non-animal classes from CIFAR10 and N non-overlapping classes from CIFAR100 \cite{sun2020cgdl, yue2021zsrosr}. As more classes are considered as open set, openness is higher. Openness for each are 46.54\% and 72.78\%.
\textbf{Tiny-ImageNet}~\cite{deng2009imagenet}
Tiny-IN is a subset of ImageNet which has 200 classes.
We follow the common protocol~\cite{neal2018osrci, guo2021caps} to resize it to 32 $\times$ 32.
Afterwards, 20 classes are sampled to be used as closed set and the remaining classes as open set classes. Openness is 68.38\%.
\begingroup
\setlength{\tabcolsep}{2.7pt} 
\renewcommand{\arraystretch}{1.2} 
\begin{table*}[t]
	\centering
	{\scriptsize
	\caption{AUROC score for detecting known and unknown samples. $\dagger$ indicates the reproduced result to unify the split information. The best results are indicated in bolds.}
	\label{table_auroc}
		\begin{tabular}{l| c c c c c c}
		    \hlineB{2.5}
			Method & MNIST & SVHN & CIFAR10 & CIFAR+10 & CIFAR+50 & Tiny-IN\\
			\hlineB{2.5}
			Softmax &  0.978 & 0.886 & 0.677 & 0.816 & 0.805 & 0.577 \\
			\hline
			OpenMax & 0.981 & 0.894 & 0.695 & 0.817 & 0.796 & 0.576 \\
			G-OpenMax\cite{ge2017gopenmax} &  0.984 & 0.896 & 0.675 & 0.827 & 0.819 & 0.580 \\
			OSRCI~\cite{neal2018osrci}& 0.988$_{\pm0.004}$ & 0.91$_{\pm0.01}$ & 0.699$_{\pm0.038}$ & 0.838 & 0.827 & 0.586 \\
			CROSR~\cite{yoshihashi2019crosr}& 0.991$_{\pm0.004}$ & 0.899$_{\pm0.018}$ & - & - & - & 0.589 \\
			C2AE~\cite{oza2019c2ae} &  - & 0.892$_{\pm0.013}$ & 0.711$_{\pm0.008}$ & 0.810$_{\pm0.005}$ & 0.803$_{\pm0.000}$ & 0.581$_{\pm0.019}$ \\
			GFROSR\cite{perera2020gdfr}& - & 0.955$_{\pm0.018}$ & 0.831$_{\pm0.039}$ & - & - & 0.657$_{\pm0.012}$ \\
			CGDL~\cite{sun2020cgdl}& 0.977$_{\pm0.008}$ & 0.896$_{\pm0.023}$ & 0.681$_{\pm0.029}$ & 0.794$_{\pm0.013}$ & 0.794$_{\pm0.003}$ & 0.653$_{\pm0.002}$ \\
			RPL~\cite{chen2020rpl}& 0.917$_{\pm0.006}$ & 0.931$_{\pm0.014}$ & 0.784$_{\pm0.025}$ & 0.885$_{\pm0.019}$ & 0.881$_{\pm0.014}$ & 0.711$_{\pm0.026}$ \\
			PROSER~\cite{zhou2021proser}$\dagger$ & 0.964$_{\pm0.019}$ & 0.930$_{\pm0.005}$ & 0.801$_{\pm0.031}$ & 0.898$_{\pm0.015}$ & 0.881$_{\pm0.003}$ & 0.684$_{\pm0.029}$ \\
			ARPL+cs~\cite{chen2021arpl}$\dagger$ & 0.991$_{\pm0.004}$ &0.946$_{\pm0.005}$& 0.819$_{\pm0.029}$ & 0.904$_{\pm0.002}$& 0.901$_{\pm0.002}$& 0.710$_{\pm0.002}$ \\
            CVAECap\cite{guo2021caps} & \textbf{0.992}$_{\pm0.004}$ & \textbf{0.956}$_{\pm0.012}$ & 0.835$_{\pm0.023}$ & 0.888$_{\pm0.019}$ & 0.889$_{\pm0.017}$ & 0.715$_{\pm0.018}$ \\
			\hline
			DIAS \textbf{(Ours)} & \textbf{0.992}$_{\pm0.004}$ & 0.943$_{\pm0.008}$ & \textbf{0.850}$_{\pm0.022}$ & \textbf{0.920}$_{\pm0.011}$ & \textbf{0.916}$_{\pm0.007}$& \textbf{0.731}$_{\pm0.015}$ \\
			\hline
            \hlineB{2.5}
		\end{tabular}
	}
\end{table*}
\endgroup
\begingroup
\setlength{\tabcolsep}{6pt} 
\renewcommand{\arraystretch}{1} 
\begin{table}[t]
	\centering

	{
		\caption{Comparison of accuracy for closed set classes between baseline and DIAS.
    }
	\label{table_acc}
	\scriptsize
		\begin{tabular}{l | c c c c c c}
		    \hlineB{2.5}
			Method & MNIST & SVHN & CIFAR10 & CIFAR+ & Tiny-IN \\
			\hlineB{2.5}
			Softmax & \textbf{0.997} & 0.966 & 0.934 & 0.960& 0.653\\
			\hline
			DIAS \textbf{(Ours)} & \textbf{0.997} & \textbf{0.970} & \textbf{0.947} & \textbf{0.964} & \textbf{0.700} \\
            \hlineB{2.5}
		\end{tabular}
	}
\end{table}
\endgroup
\begingroup
\setlength{\tabcolsep}{4.0pt} 
\renewcommand{\arraystretch}{0.8} 
\begin{table}[t]
	\centering
	{
		\caption{
    Average of macro-averaged F1-scores in five splits. 
	We adopt the protocol from GCM-CF~\cite{yue2021zsrosr}. $\dagger$ indicates the reproduced performance with official code.}
	\label{table_f1}
		\begin{tabular}{l | c c c c c c c}
		    \hlineB{2.5}
			Method & MNIST & SVHN & CIFAR10 & CIFAR+10 & CIFAR+50 \\
			\hlineB{2.5}
			Softmax & 0.767 & 0.762 & 0.704 & 0.778 & 0.660 \\
			\hline
			Openmax & 0.859 & 0.780 & 0.714 & 0.787 & 0.677 \\
			CGDL~\cite{sun2020cgdl} & 0.890 & 0.763 & 0.710 & 0.779 & 0.710 \\
            GCM-CF~\cite{yue2021zsrosr} & 0.914 & 0.793 & 0.726 & 0.794 & 0.746 \\
            ARPL+cs~\cite{chen2021arpl} $\dagger$ & ${0.951}_{\pm{0.009}}$ & ${0.857}_{\pm{0.008}}$ & ${0.753}_{\pm{0.033}}$ & ${0.827}_{\pm{0.010}}$ & ${0.753}_{\pm{0.001}}$ \\
			\hline
		  DIAS (\textbf{Ours}) & ${\textbf{0.953}}_{\pm{0.015}}$ & ${\textbf{0.880}}_{\pm{0.010}}$ & ${\textbf{0.809}}_{\pm{0.026}}$ & ${\textbf{0.859}}_{\pm{0.010}}$ & ${\textbf{0.829}}_{\pm{0.006}}$ \\ 
			\hline
            \hlineB{2.5}
		\end{tabular}
	}
\end{table}
\endgroup

\subsection{Experimental Results}
OSR performances are in Tab.~\ref{table_auroc} and Tab.~\ref{table_f1}.
Most of the baseline results in Tab.~\ref{table_auroc} are taken from the \cite{guo2021caps} where they reproduced papers' performances with the same configuration for a fair comparison.
The scores of DIAS, PROSER~\cite{zhou2021proser}, and ARPL+cs~\cite{chen2021arpl} are evaluated based on the same protocol with \cite{guo2021caps} on the split publicized by \cite{guo2021caps, neal2018osrci}. 
We use their hyperparameters except the model architecture in PROSER to unify the backbone.
Note that some papers cannot be compared under the same codebase due to non-reproducible results (split information nor the codes are not publicized)~\cite{zhang2020hybrid} and the contrasting assumption in the existence of open set data~\cite{kong2021opengan}.
As shown, our method achieves significant improvements over the state-of-the-art techniques in CIFAR10, CIFAR+10, CIFAR+50, and tiny-ImageNet datasets, and shows comparable results in digit datasets where the performances are almost saturated.
Also, Tab.~\ref{table_acc} shows that DIAS improves the accuracy of the closed set along with its capability of detecting unknowns.


\begingroup
\setlength{\tabcolsep}{6pt} 
\renewcommand{\arraystretch}{0.7} 
\begin{table}[t]
	\centering
{
	\caption{Macro-averaged F1-scores on the MNIST with three other datasets as unknown.}
	\label{table_mnist}
    \begin{tabular}{l|c c c}
    \hlineB{2.5}
    Method     & Omniglot & MNIST-noise & Noise \\ \hlineB{2.5}
    Softmax    & 0.595                         & 0.801       & 0.829 \\
    Openmax~\cite{bendale2016openmax}    & 0.780                         & 0.816       & 0.826 \\
    CROSR~\cite{yoshihashi2019crosr}      & 0.793                         & 0.827       & 0.826 \\
    CGDL~\cite{sun2020cgdl}       & 0.850                         & 0.887       & 0.859 \\
    PROSER~\cite{zhou2021proser}     & 0.862                         & 0.874       & 0.882 \\
    CVAECapOSR~\cite{guo2021caps} & 0.971                         & \textbf{0.982}       & 0.982 \\ \hline
    DIAS \textbf{(Ours)}       & \textbf{0.989}                & \textbf{0.982}       & \textbf{0.989}       \\ \hlineB{2.5}
    \end{tabular}

}
\end{table}
\endgroup


We discussed that DIAS establishes a standard to determine a proper threshold for unknowns in Sec.~\ref{sec:inference}.
In Tab.~\ref{table_f1}, we validate such claim that
DIAS does not require expensive and complex tuning for the threshold search so thus it is much more practical than previous works. 
For a fair comparison, baseline results are from \cite{yue2021zsrosr}, and ARPL+cs and ours are tested on the same split.
The results show that DIAS consistently outperforms the baselines with noticeable margins.
For the threshold, we find $\epsilon$ in Eq.~\ref{eq:inference} works well when set to -0.05 for all experiments.

As previously did in \cite{sun2020cgdl, guo2021caps, zhou2021proser}, we conduct an additional open set detection experiment.
Briefly, we train the classifier on MNIST and evaluate on Omniglot~\cite{lake2015omniglot}, MNIST-Noise, and Noise.
Omniglot is an alphabet dataset 
of 1623 handwritten characters from 50 writing systems,
and Noise is a synthesized dataset where each pixel is sampled from a gaussian distribution. MNIST-Noise is noise-embedded MNIST dataset.
We sample 10000 examples from each dataset since MNIST contains 10000 instances. The macro F-score between ten digit classes and open set classes are measured to compare performances.
The experimental results are reported in Tab.~\ref{table_mnist}. Although the performances on all three datasets are almost saturated, DIAS provides competitive results with state-of-the-art methods.

\subsection{Ablation study and Further analysis}
\label{sec:ablation}
\begingroup
\setlength{\tabcolsep}{6pt} 
\renewcommand{\arraystretch}{0.7} 
\begin{table}[t]
	\centering
    {
    \caption{Ablation study on varying difficulties of fake examples. We report the AUROC scores on CIFAR100 dataset against varying openness.}
    \label{table_ablation_layer}
    \begin{tabular}{c|c|c|lllll}
    \hlineB{2.5}
    Copycat & GAN & Copycat  & \multicolumn{4}{c}{Openness (\%)}                    \\ \cline{4-7}
    Hard & Moderate & Easy & 22.54 & 29.29 & 55.28 & 68.38 \\ \hlineB{2.5}
    -  & - & - & 72.28 & 71.72 & 78.85 & 77.34 \\
    -  & - & \cmark & 72.45 & 72.51 & 78.67 & 78.36 \\
    -  & \cmark & - & 73.69 & 72.46 & 79.20 & 81.13 \\
    \cmark  & - & - & 76.59 & 74.82 & 81.05 & 80.91 \\
    \cmark  & - & \cmark & 76.79 & 74.92 & 81.06 & 81.56 \\
    \cmark  & \cmark & \cmark & \textbf{76.92} & \textbf{75.55} & \textbf{81.59} & \textbf{83.95} \\ \hlineB{2.5}
    \end{tabular}
    }
\end{table}
\endgroup
\begingroup
\setlength{\tabcolsep}{6pt} 
\renewcommand{\arraystretch}{0.8} 
\begin{table}[t]
	\centering
    {
    \caption{Ablation study on each generator in DIAS with F1-score.}
    \label{table_ablation_generator}
    \begin{tabular}{cc|lllll}
    \hlineB{2.5}
    Copycat & GAN & MNIST & SVHN & C10 & C+10 & C+50\\
    \hlineB{2.5}
    - & - & 0.767 & 0.762 & 0.704 & 0.778 & 0.660 \\
    -  & \cmark & 0.926 & 0.840 & 0.777 & 0.850 & 0.775\\
    \cmark & - & 0.948 & 0.860 & 0.788 & 0.846 & 0.816\\
    \cmark  & \cmark & \textbf{0.953} & \textbf{0.880} & \textbf{0.809} & \textbf{0.859} & \textbf{0.829}\\ \hlineB{2.5}
    \end{tabular}
    }
\end{table}
\endgroup
\begingroup
\setlength{\tabcolsep}{6.0pt} 
\renewcommand{\arraystretch}{0.8} 
\begin{table}[t]
	\centering
	{
		\caption{
    Validity of the inherent threshold in DIAS. For DIAS (*), we searched the best threshold for identifying unknowns in DIAS.}
	\small
	\label{table_ablation_threshold}
		\begin{tabular}{l | c c c c c c c}
		    \hlineB{2.5}
			Method & MNIST & SVHN & CIFAR10 & CIFAR+10 & CIFAR+50 \\
			\hlineB{2.5}
		  DIAS & ${0.953}_{\pm{0.015}}$ & ${0.880}_{\pm{0.010}}$ & ${0.809}_{\pm{0.026}}$ & ${0.859}_{\pm{0.010}}$ & ${0.829}_{\pm{0.006}}$ \\
		  DIAS (*) & ${0.970}_{\pm{0.004}}$ & ${0.883}_{\pm{0.009}}$ & ${0.809}_{\pm{0.026}}$ & ${0.861}_{\pm{0.009}}$ & ${0.833}_{\pm{0.005}}$ \\

			\hline
            \hlineB{2.5}
		\end{tabular}
	}

\end{table}
\endgroup
\textbf{Effect of varying difficulty levels.}
In Tab.~\ref{table_ablation_layer}, we summarize the ablation analysis on varying difficulties of fake samples.
As reported, our hard-difficulty fakes have highest contribution to improve the OSR performance.
It validates that the detailed calibration of decision boundaries by facing with hard-difficulty fake examples is significantly helpful to enhance the model robustness toward unseen samples.
For the relatively small improvement brought by the easy-difficulty examples, we think that the supervised models are already quite robust against such easy cases.
More importantly, by utilizing all the difficulty-level samples for the simulation, DIAS boosts the AUROC at various openness configurations.
On top of this, we also evaluate each generator in Tab.~\ref{table_ablation_generator} with F1-score to show their relative importance.
%
Note that, we search for the best threshold 
by Eq.~\ref{eq:inference} to distinguish open set with F1-score for approaches without the Copycat, while it occurs by itself for DIAS when processing the hard fakes of Copycat.
%
Results demonstrate that both our generators are suitable for simulating open set instances and also validate the unique advantages of Copycat; significantly improving the performance with its inherent threshold for identifying unknowns.

\begin{figure}[t]
    \centering
    \includegraphics[width=1\textwidth]{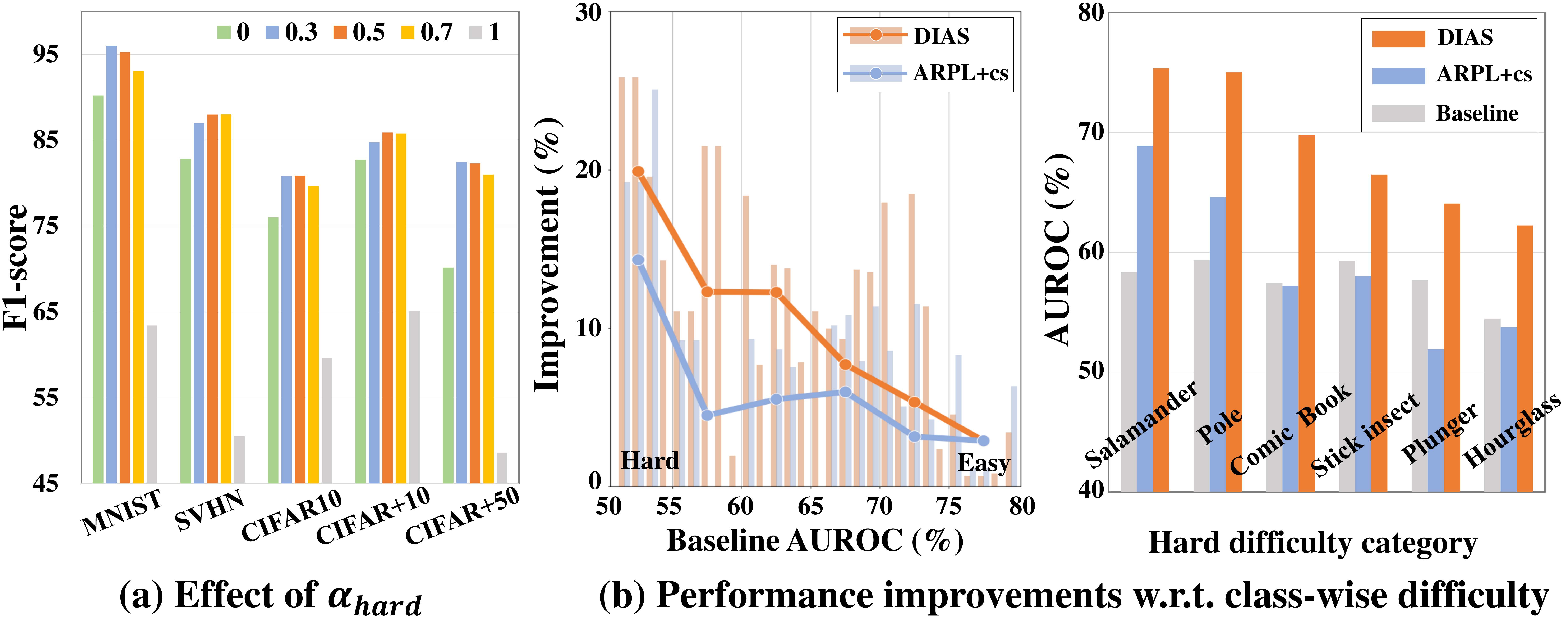}
    \caption{\textbf{(a)} Effect of $\alpha_{hard}$. Results demonstrate that hard fake samples significantly contribute to calibrate the decision boundary. Regarding these as either known (green) or unknown (gray) decreases the performances.
    \textbf{(b)} Performance improvements over the baseline Softmax classifier w.r.t. the class-wise difficulty.
    \textbf{(b-Left)} The difficulty of open classes are determined by the AUROC scores of the Softmax classifier (x-axis). 
    Thus, the lower AUROC scores, the harder the classes are, and placed on the left side.
    The bars in the graph show the class-wise improvements from the baseline Softmax classifier in order of the difficulty level, while the curves represent the average improvements over classes within 5\% intervals on the AUROC scores of the Softmax classifier.
    As shown, DIAS is more effective in distinguishing harder open classes.
    \textbf{(b-Right)} Comparison of class-specific AUROC scores on harder open classes. Such results validate the effectiveness of DIAS for identifying open set classes which our baseline classifiers find difficult.
    }
	\label{fig_smoothing_ablation}
\end{figure}

\textbf{Effect of smoothing ratio.}
Since $\alpha_{hard}$ is an important parameter, we study its impact in Fig.~\ref{fig_smoothing_ablation}~(a).
Intuitively, the target label for the fake samples become uniform distribution when $\alpha_{hard}$ is 1, while it becomes one-hot label with $\alpha_{hard}$ of 0.
The tendency observed from the gray bar with $\alpha_{hard}=1$, forcing the hard-difficulty samples to have uniformly distributed class probability drastically degrades the performance.
On the other hand, the performance is also dropped if we treat the fake samples as known classes with $\alpha_{hard}=0$, since in such case the hard-difficulty samples are no longer utilized to calibrate within the class decision boundaries.
Excepting those extreme cases, we observe that DIAS has low sensitivity to the choice of the hyperparameter $\alpha_{hard}$.
Note that, all the results reported in Tab.~\ref{table_f1} is produced with $\alpha_{hard}$ of 0.5.


\textbf{Validity of the inherent threshold}
 is confirmed in Tab.~\ref{table_ablation_threshold}.
Although the optimal threshold may differ between datasets, only small gaps between the best and inherent thresholds verify that smoothed probability for hard fake examples is suitable to be an adequate threshold because there is no searching cost.

\textbf{Further analysis}
We conducted an in-depth analysis on Tiny-ImageNet to explore why the proposed method is effective in detecting unknowns and how effective it is on each class and each difficulty group.
For this study, we assume that difficulty levels are only varying across classes. In other words, we do not consider the instance-wise difficulty in this experiment.
Specifically, we utilize class-wise AUROC score of the Softmax classifier to determine the difficulty.
The class-wise improvements of the OSR is reported in Fig.~\ref{fig_smoothing_ablation} (b) with comparison against the most recent simulation method~\cite{chen2021arpl}. 
Those results validate the merits of proposed DIAS especially in distinguishing confusing known and unknown instances, while it provides improvements across all levels of difficulties.

\begin{figure}[t]
    \centering
    \includegraphics[width=0.9\linewidth]{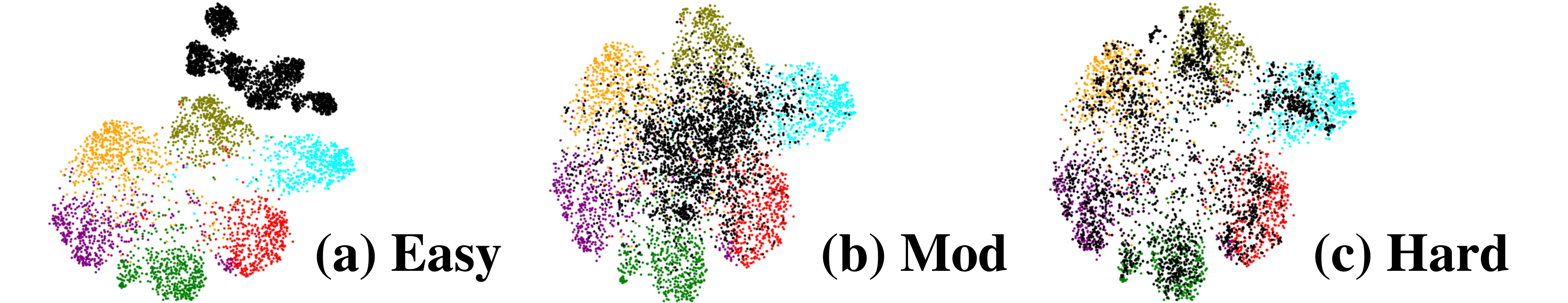}
    \caption{
      t-SNE of fake 
  distributions (Black). 
  As the difficulty level gets higher, black dots are harder to be separated.
    }
    \label{fig:tsne}
\end{figure}

To understand how DIAS enables better separation in all difficulty levels, we examine our generators. 
Specifically, we visualized the feature space of the Softmax classifier with fake samples from our generators on CIFAR10.
In Fig.~\ref{fig:tsne}, six colored clusters other than black correspond to each closed set class. We find that our generators are actually generating diverse fakes as we intentionally designed.
Easy fakes are embedded out of class clusters (Fig.~\ref{fig:tsne} (a)), moderate ones are partially mixed with known samples (Fig.~\ref{fig:tsne} (b)), and finally, hard-difficulty examples significantly overlie on top of the knowns (Fig.~\ref{fig:tsne} (c)).
Hence, as our virtual open set covers broader range of difficulty levels in open set, DIAS produced better results across all difficulty levels.

In addition, one may ask how the Copycat is able to generate more confusing fakes than GAN.
This is because the generator in GAN learn to generate fakes that share features with knowns in the viewpoint of the discriminator, while the difficulty levels depend on the viewpoint of the classifier. 
As the Copycat learns the classifier's perspective iteratively, the Copycat is equipped with the strength to generate confusing fake instances from the classifier's viewpoint.
AUROC and W.D from Fig.~\ref{figure_motivation} and visualized feature maps in Fig.~\ref{fig:tsne} further support the choice of setting the Copycat as the most difficult-fake instance generator.


\section{Conclusion}
\label{sec:conclusion}
OSR assumes numerous objects are present that do not belong to learned classes. 
When classifiers are facing with these, they misclassify them into one of the known categories, often with high confidence. 
To prepare the classifier for handling unknowns, there have been works to simulate virtual open instances.
However, these works only considered unknowns as one set.
We claim that considering various levels of difficulty in OSR is an untapped question to be studied.
To this end, we proposed the Difficulty-Aware Simulator to simulate open set with fakes at various difficulty levels.
Also, we introduced the Copycat and the GAN-based generator in the classifier's perspective for preparing adequate fake samples for classifier tuning.
Extensive experiments demonstrate that our proposed DIAS significantly improves the robustness of the classifier toward open set instances.

\vspace{5pt}
\noindent\textbf{Acknowledgements.} 
This work was supported in part by MSIT/IITP (No. 2022-0-00680, 2020-0-00973, 2020-0-01821, and 2019-0-00421), MCST/KOCCA (No. R2020070002), and MSIT\&KNPA/KIPoT (Police Lab 2.0, No. 210121M06).
\bibliographystyle{splncs04}
\bibliography{egbib}

\begin{thebibliography}{10}
\providecommand{\url}[1]{\texttt{#1}}
\providecommand{\urlprefix}{URL }
\providecommand{\doi}[1]{https://doi.org/#1}

\bibitem{arora2017generalization}
Arora, S., Ge, R., Liang, Y., Ma, T., Zhang, Y.: Generalization and equilibrium
  in generative adversarial nets (gans). In: International Conference on
  Machine Learning. PMLR (2017)

\bibitem{bendale2015towards}
Bendale, A., Boult, T.: Towards open world recognition. In: Proceedings of the
  IEEE conference on computer vision and pattern recognition (2015)

\bibitem{bendale2016openmax}
Bendale, A., Boult, T.E.: Towards open set deep networks. In: Proceedings of
  the IEEE conference on computer vision and pattern recognition (2016)

\bibitem{chen2021arpl}
Chen, G., Peng, P., Wang, X., Tian, Y.: Adversarial reciprocal points learning
  for open set recognition. IEEE Transactions on Pattern Analysis and Machine
  Intelligence (TPAMI)  (2021). \doi{10.1109/TPAMI.2021.3106743}

\bibitem{chen2020rpl}
Chen, G., Qiao, L., Shi, Y., Peng, P., Li, J., Huang, T., Pu, S., Tian, Y.:
  Learning open set network with discriminative reciprocal points. In: Computer
  Vision--ECCV 2020: 16th European Conference, Glasgow, UK, August 23--28,
  2020, Proceedings, Part III 16. Springer (2020)

\bibitem{chen2021distilling}
Chen, P., Liu, S., Zhao, H., Jia, J.: Distilling knowledge via knowledge
  review. In: Proceedings of the IEEE/CVF Conference on Computer Vision and
  Pattern Recognition (2021)

\bibitem{deng2009imagenet}
Deng, J., Dong, W., Socher, R., Li, L.J., Li, K., Fei-Fei, L.: Imagenet: A
  large-scale hierarchical image database. In: 2009 IEEE conference on computer
  vision and pattern recognition. Ieee (2009)

\bibitem{ge2017gopenmax}
Ge, Z., Demyanov, S., Chen, Z., Garnavi, R.: Generative openmax for multi-class
  open set classification. In: Proceedings of the British Machine Vision
  Conference (BMVC) (2017)

\bibitem{geirhos2020shortcut}
Geirhos, R., Jacobsen, J.H., Michaelis, C., Zemel, R., Brendel, W., Bethge, M.,
  Wichmann, F.A.: Shortcut learning in deep neural networks. Nature Machine
  Intelligence  (2020)

\bibitem{girish2021towards}
Girish, S., Suri, S., Rambhatla, S.S., Shrivastava, A.: Towards discovery and
  attribution of open-world gan generated images. In: Proceedings of the
  IEEE/CVF International Conference on Computer Vision. pp. 14094--14103 (2021)

\bibitem{goodfellow2014gan}
Goodfellow, I., Pouget-Abadie, J., Mirza, M., Xu, B., Warde-Farley, D., Ozair,
  S., Courville, A., Bengio, Y.: Generative adversarial nets. Advances in
  neural information processing systems  (2014)

\bibitem{guo2017calibration}
Guo, C., Pleiss, G., Sun, Y., Weinberger, K.Q.: On calibration of modern neural
  networks. In: International Conference on Machine Learning. PMLR (2017)

\bibitem{guo2021caps}
Guo, Y., Camporese, G., Yang, W., Sperduti, A., Ballan, L.: Conditional
  variational capsule network for open set recognition. Proceedings of the IEEE
  international conference on computer vision (ICCV)  (2021)

\bibitem{he2016deep}
He, K., Zhang, X., Ren, S., Sun, J.: Deep residual learning for image
  recognition. In: Proceedings of the IEEE conference on computer vision and
  pattern recognition (2016)

\bibitem{he2016resnet}
He, K., Zhang, X., Ren, S., Sun, J.: Deep residual learning for image
  recognition. In: Proceedings of the IEEE conference on computer vision and
  pattern recognition (2016)

\bibitem{hendrycks2016baseline}
Hendrycks, D., Gimpel, K.: A baseline for detecting misclassified and
  out-of-distribution examples in neural networks. International Conference on
  Learning Representations (ICLR)  (2017)

\bibitem{hinton2015distilling}
Hinton, G., Vinyals, O., Dean, J.: Distilling the knowledge in a neural
  network. In: NIPS Workshop on Deep Learning and Representation Learning
  Workshop (2015)

\bibitem{godin}
Hsu, Y.C., Shen, Y., Jin, H., Kira, Z.: Generalized odin: Detecting
  out-of-distribution image without learning from out-of-distribution data. In:
  Proceedings of the IEEE/CVF Conference on Computer Vision and Pattern
  Recognition (2020)

\bibitem{huang2017densenet}
Huang, G., Liu, Z., Van Der~Maaten, L., Weinberger, K.Q.: Densely connected
  convolutional networks. In: Proceedings of the IEEE conference on computer
  vision and pattern recognition (2017)

\bibitem{jain2014multi}
Jain, L.P., Scheirer, W.J., Boult, T.E.: Multi-class open set recognition using
  probability of inclusion. In: European Conference on Computer Vision.
  Springer (2014)

\bibitem{junior2017nearest}
J{\'u}nior, P.R.M., De~Souza, R.M., Werneck, R.d.O., Stein, B.V., Pazinato,
  D.V., de~Almeida, W.R., Penatti, O.A., Torres, R.d.S., Rocha, A.: Nearest
  neighbors distance ratio open-set classifier. Machine Learning  (2017)

\bibitem{kong2021opengan}
Kong, S., Ramanan, D.: Opengan: Open-set recognition via open data generation.
  In: ICCV (2021)

\bibitem{krizhevsky2009learning}
Krizhevsky, A., Hinton, G., et~al.: Learning multiple layers of features from
  tiny images  (2009)

\bibitem{krizhevsky2012imagenet}
Krizhevsky, A., Sutskever, I., Hinton, G.E.: Imagenet classification with deep
  convolutional neural networks. Advances in neural information processing
  systems  (2012)

\bibitem{lake2015omniglot}
Lake, B.M., Salakhutdinov, R., Tenenbaum, J.B.: Human-level concept learning
  through probabilistic program induction. Science  (2015)

\bibitem{lecun1998mnist}
LeCun, Y., Bottou, L., Bengio, Y., Haffner, P.: Gradient-based learning applied
  to document recognition. Proceedings of the IEEE  (1998)

\bibitem{lee2017training}
Lee, K., Lee, H., Lee, K., Shin, J.: Training confidence-calibrated classifiers
  for detecting out-of-distribution samples. International Conference on
  Learning Representations (ICLR)  (2018)

\bibitem{neal2018osrci}
Neal, L., Olson, M., Fern, X., Wong, W.K., Li, F.: Open set learning with
  counterfactual images. In: Proceedings of the European Conference on Computer
  Vision (ECCV) (2018)

\bibitem{netzer2011reading}
Netzer, Y., Wang, T., Coates, A., Bissacco, A., Wu, B., Ng, A.Y.: Reading
  digits in natural images with unsupervised feature learning. In: NIPS
  Workshop on Deep Learning and Unsupervised Feature Learning (2011)

\bibitem{nguyen2015deep}
Nguyen, A., Yosinski, J., Clune, J.: Deep neural networks are easily fooled:
  High confidence predictions for unrecognizable images. In: Proceedings of the
  IEEE conference on computer vision and pattern recognition (2015)

\bibitem{oza2019c2ae}
Oza, P., Patel, V.M.: C2ae: Class conditioned auto-encoder for open-set
  recognition. In: Proceedings of the IEEE/CVF Conference on Computer Vision
  and Pattern Recognition (2019)

\bibitem{perera2020gdfr}
Perera, P., Morariu, V.I., Jain, R., Manjunatha, V., Wigington, C., Ordonez,
  V., Patel, V.M.: Generative-discriminative feature representations for
  open-set recognition. In: Proceedings of the IEEE/CVF Conference on Computer
  Vision and Pattern Recognition (2020)

\bibitem{roady2020difsim}
Roady, R., Hayes, T.L., Kemker, R., Gonzales, A., Kanan, C.: Are open set
  classification methods effective on large-scale datasets? Plos one  (2020)

\bibitem{russakovsky2015imagenet}
Russakovsky, O., Deng, J., Su, H., Krause, J., Satheesh, S., Ma, S., Huang, Z.,
  Karpathy, A., Khosla, A., Bernstein, M., et~al.: Imagenet large scale visual
  recognition challenge. International journal of computer vision  (2015)

\bibitem{scheirer2014probability}
Scheirer, W.J., Jain, L.P., Boult, T.E.: Probability models for open set
  recognition. IEEE transactions on pattern analysis and machine intelligence
  (2014)

\bibitem{scheirer2012toward}
Scheirer, W.J., de~Rezende~Rocha, A., Sapkota, A., Boult, T.E.: Toward open set
  recognition. IEEE transactions on pattern analysis and machine intelligence
  (2012)

\bibitem{shu2017doc}
Shu, L., Xu, H., Liu, B.: Doc: Deep open classification of text documents.
  Proceedings of the 2017 Conference on Empirical Methods in Natural Language
  Processing, (EMNLP)  (2017)

\bibitem{sun2020cgdl}
Sun, X., Yang, Z., Zhang, C., Ling, K.V., Peng, G.: Conditional gaussian
  distribution learning for open set recognition. In: Proceedings of the
  IEEE/CVF Conference on Computer Vision and Pattern Recognition (CVPR) (2020)

\bibitem{tudor2016hard}
Tudor~Ionescu, R., Alexe, B., Leordeanu, M., Popescu, M., Papadopoulos, D.P.,
  Ferrari, V.: How hard can it be? estimating the difficulty of visual search
  in an image. In: Proceedings of the IEEE Conference on Computer Vision and
  Pattern Recognition (2016)

\bibitem{verma2019manifold}
Verma, V., Lamb, A., Beckham, C., Najafi, A., Mitliagkas, I., Lopez-Paz, D.,
  Bengio, Y.: Manifold mixup: Better representations by interpolating hidden
  states. In: International Conference on Machine Learning. PMLR (2019)

\bibitem{yim2017gift}
Yim, J., Joo, D., Bae, J., Kim, J.: A gift from knowledge distillation: Fast
  optimization, network minimization and transfer learning. In: Proceedings of
  the IEEE Conference on Computer Vision and Pattern Recognition (2017)

\bibitem{yoshihashi2019crosr}
Yoshihashi, R., Shao, W., Kawakami, R., You, S., Iida, M., Naemura, T.:
  Classification-reconstruction learning for open-set recognition. In:
  Proceedings of the IEEE/CVF Conference on Computer Vision and Pattern
  Recognition (2019)

\bibitem{yue2021zsrosr}
Yue, Z., Wang, T., Sun, Q., Hua, X.S., Zhang, H.: Counterfactual zero-shot and
  open-set visual recognition. In: Proceedings of the IEEE/CVF Conference on
  Computer Vision and Pattern Recognition (CVPR) (2021)

\bibitem{zagoruyko2016paying}
Zagoruyko, S., Komodakis, N.: Paying more attention to attention: Improving the
  performance of convolutional neural networks via attention transfer.
  International Conference on Learning Representations (ICLR)  (2017)

\bibitem{zhang2016sparse}
Zhang, H., Patel, V.M.: Sparse representation-based open set recognition. IEEE
  transactions on pattern analysis and machine intelligence  (2016)

\bibitem{zhang2020hybrid}
Zhang, H., Li, A., Guo, J., Guo, Y.: Hybrid models for open set recognition.
  In: European Conference on Computer Vision. Springer (2020)

\bibitem{zhou2021proser}
Zhou, D.W., Ye, H.J., Zhan, D.C.: Learning placeholders for open-set
  recognition. In: Proceedings of the IEEE/CVF Conference on Computer Vision
  and Pattern Recognition (2021)

\end{thebibliography}
\newpage
\section{Split Information}
As we elaborated in the main paper, we adopted the protocols from \cite{guo2021caps} and \cite{yue2021zsrosr} for evaluations with AUROC and F1-score, respectively.
To further encourage the fair comparison, we publicize the split details.
Specifically, we enumerate categories that are used for closed-set in Tab.~\ref{table_f1_split} and Tab.~\ref{table_auroc_split} for measuring F1-score and AUROC, respectively.
Note that for CIFAR+, we show the categories of open-set classes since CIFAR+ experiments utilize the non-animal classes in CIFAR10 dataset, i.e., airplain, automobile, ship, and truck, as the closed-set.
We sincerely hope future works use pre-defined standard split information to prevent confusion in understanding the effectiveness of their methods and for a fair comparison.

\begingroup
\setlength{\tabcolsep}{6pt} 
\renewcommand{\arraystretch}{1.1} 
\begin{table*}[ht]
    \caption{
    Data splits for Tab. 3 in the main paper. This split information is used for measuring F1-scores. The numbers in the table represent the class indices for closed set except CIFAR+ cases. For CIFAR+ experiments, we provide open-set class indices, since animal classes are utilized for closed set.
	}
	\label{table_f1_split}
\scriptsize
\begin{tabularx}{\textwidth}{|l|X|X|X|X|X|}
\hlineB{2.5}
\multicolumn{6}{|c|}{\textbf{F1 Split Information}} \\
\hlineB{2.5}
\multicolumn{1}{|c|}{\textbf{}} & \textbf{0} & \textbf{1}& \textbf{2}   & \textbf{3}& \textbf{4}    \\ \hlineB{2.5}
MNIST & 2, 3, 4, 6, 7, 8    & 0, 1, 4, 6, 7, 9  & 1, 2, 4, 6, 7, 8 & 1, 3, 4, 6, 7, 8 & 1, 2, 3, 5, 7, 8 \\
\hline
SVHN  & 2, 3, 4, 6, 7, 8    & 0, 1, 4, 6, 7, 9  & 1, 2, 4, 6, 7, 8 & 1, 3, 4, 6, 7, 8 & 1, 2, 3, 5, 7, 8 \\\hline
CIFAR10    & 2, 3, 4, 6, 7, 8    & 0, 1, 4, 6, 7, 9  & 1, 2, 4, 6, 7, 8 & 1, 3, 4, 6, 7, 8 & 1, 2, 3, 5, 7, 8 \\\hline
CIFAR+10   & 27, 46, 98, 38, 72, 31, 36, 66, 3, 97 & 98, 46, 14, 1, 7, 73, 3, 79, 93, 11 & 79, 98, 67, 7, 77, 42, 36, 65, 26, 64  & 46, 77, 29, 24, 65, 66, 79, 21, 1, 95 & 21, 95, 64, 55, 50, 24, 93, 75, 27, 36  \\\hline
CIFAR+50   & 27, 46, 98, 38, 72, 31, 36, 66, 3, 97, 75, 67, 42, 32, 14, 93, 6, 88, 11, 1, 44, 35, 73, 19, 18, 78, 15, 4, 50, 65, 64, 55, 30, 80, 26, 2, 7, 34, 79, 43, 74, 29, 45, 91, 37, 99, 95, 63, 24, 21 &    98, 46, 14, 1, 7, 73, 3, 79, 93, 11, 37, 29, 2, 74, 91, 77, 55, 50, 18, 80, 63, 67, 4, 45, 95, 30, 75, 97, 88, 36, 31, 27, 65, 32, 43, 72, 6, 26, 15, 42, 19, 34, 38, 66, 35, 21, 24, 99, 78, 44 &    79, 98, 67, 7, 77, 42, 36, 65, 26, 64, 66, 73, 75, 3, 32, 14, 35, 6, 24, 21, 55, 34, 30, 43, 93, 38, 19, 99, 72, 97, 78, 18, 31, 63, 29, 74, 91, 4, 27, 46, 2, 88, 45, 15, 11, 1, 95, 50, 80, 44 &       46, 77, 29, 24, 65, 66, 79, 21, 1, 95, 36, 88, 27, 99, 67, 19, 75, 42, 2, 73, 32, 98, 72, 97, 78, 11, 14, 74, 50, 37, 26, 64, 44, 30, 31, 18, 38, 4, 35, 80, 45, 63, 93, 34, 3, 43, 6, 55, 91, 15 &     21, 95, 64, 55, 50, 24, 93, 75, 27, 36, 73, 63, 19, 98, 46, 1, 15, 72, 42, 78, 77, 29, 74, 30, 14, 38, 80, 45, 4, 26, 31, 11, 97, 7, 66, 65, 99, 34, 6, 18, 44, 3, 35, 88, 43, 91, 32, 67, 37, 79 \\\hlineB{2.5}
\end{tabularx}
\end{table*}
\endgroup
\begingroup
\setlength{\tabcolsep}{6pt} 
\renewcommand{\arraystretch}{1.1} 
\begin{table*}[ht]
    \caption{
    Data splits for Tab. 1 in the main paper. This split information is used for measuring AUROC scores. The numbers in the table represent the class indices for closed set except CIFAR+ cases. For CIFAR+ experiments, we provide open-set class indices, since animal classes are utilized for closed set.
	}
	\label{table_auroc_split}
\scriptsize
\begin{tabularx}{\textwidth}{|l|X|X|X|X|X|}
\hlineB{2.5}
\multicolumn{6}{|c|}{\textbf{AUROC Split Information}} \\
\hlineB{2.5}
\multicolumn{1}{|c|}{\textbf{}} & \textbf{0} & \textbf{1}& \textbf{2}   & \textbf{3}& \textbf{4}    \\ \hlineB{2.5}
MNIST & 0, 1, 2, 4, 5, 9    & 0, 3, 5, 7, 8, 9   & 0, 1, 5, 6, 7, 8 & 3, 4, 5, 7, 8, 9   & 0, 1, 2, 3, 7, 8   \\
\hline
SVHN  & 0, 1, 2, 4, 5, 9    & 0, 3, 5, 7, 8, 9   & 0, 1, 5, 6, 7, 8 & 3, 4, 5, 7, 8, 9   & 0, 1, 2, 3, 7, 8   \\\hline
CIFAR10    & 0, 1, 2, 4, 5, 9    & 0, 3, 5, 7, 8, 9   & 0, 1, 5, 6, 7, 8 & 3, 4, 5, 7, 8, 9   & 0, 1, 2, 3, 7, 8   \\\hline
CIFAR+10   & 26, 31, 34, 44, 45, 63, 65, 77, 93, 98  & 7, 11, 66, 75, 77, 93, 95, 97, 98, 99  & 2, 11, 15, 24, 32, 34, 63, 88, 93, 95& 1, 11, 38, 42, 44, 45, 63, 64, 66, 67  & 3, 15, 19, 21, 42, 46, 66, 72, 78, 98  \\\hline
CIFAR+50   & 1, 2, 7, 9, 10, 12, 15, 18, 21, 23, 26, 30, 32, 33, 34, 36, 37, 39, 40, 42, 44, 45, 46, 47, 49, 50, 51, 52, 55, 56, 59, 60, 61, 63, 65, 66, 70, 72, 73, 74, 76, 78, 80, 83, 87, 91, 92, 96, 98, 99 & 0, 2, 4, 5, 9, 12, 14, 17, 18, 20, 21, 23, 24, 25, 31, 32, 33, 35, 39, 43, 45, 49, 50, 51, 52, 54, 55, 56, 60, 64, 65, 66, 68, 70, 71, 73, 74, 77, 78, 79, 80, 82, 83, 86, 91, 93, 94, 96, 97, 98 & 0, 4, 10, 11, 12, 14, 15, 17, 18, 21, 23, 26, 27, 28, 29, 31, 32, 33, 36, 39, 40, 42, 43, 46, 47, 51, 53, 56, 57, 59, 60, 64, 66, 71, 73, 74, 75, 76, 78, 79, 80, 83, 87, 91, 92, 93, 94, 95, 96, 99 & 0, 2, 5, 6, 9, 10, 11, 12, 14, 16, 18, 19, 21, 22, 23, 26, 27, 28, 29, 31, 33, 35, 36, 37, 38, 39, 40, 43, 45, 49, 52, 56, 59, 61, 62, 63, 64, 65, 71, 74, 75, 78, 80, 82, 86, 87, 91, 93, 94, 96 & 0, 1, 4, 6, 7, 12, 15, 16, 17, 19, 20, 21, 22, 23, 26, 27, 28, 32, 39, 40, 42, 43, 44, 47, 49, 50, 52, 53, 54, 55, 56, 59, 61, 62, 63, 65, 66, 67, 68, 73, 74, 77, 82, 83, 86, 87, 93, 94, 97, 98 \\\hline
Tiny-IN   & 2, 3, 13, 30, 44, 45, 64, 66, 76, 101, 111, 121, 128, 130, 136, 158, 167, 170, 187, 193   & 4, 11, 32, 42, 51, 53, 67, 84, 87, 104, 116, 140, 144, 145, 148, 149, 155, 168, 185, 193 & 3, 9, 10, 20, 23, 28, 29, 45, 54, 74, 133, 143, 146, 147, 156, 159, 161, 170, 184, 195 & 1, 15, 17, 31, 36, 44, 66, 69, 84, 89, 102, 137, 154, 160, 170, 177, 182, 185, 195, 197  & 4, 14, 16, 33, 34, 39, 59, 69, 77, 92, 101, 103, 130, 133, 147, 161, 166, 168, 172, 173  \\ \hlineB{2.5}
\end{tabularx}
\end{table*}
\endgroup

\newpage
\section{Regularization Loss}
As we introduced in the main paper, we simply used cross entropy loss function for regularization loss, $\mathcal{L}_{reg}$.
In this section, we simply examine the influence of $\mathcal{L}_{reg}$ with two datasets: CIFAR10 and Tiny-ImageNet.
Results in Tab.~\ref{table_regratio} show that DIAS is not very sensitive to the ratio for $\mathcal{L}_{reg}$.
\begingroup
\setlength{\tabcolsep}{6pt} 
\renewcommand{\arraystretch}{1} 
\begin{table}[h]
	\vspace{-0.35cm}
	\centering
	{\small
        \begin{tabular}{c | c c c c}
        \hlineB{2.5}
        Loss Ratio & 0.1 & 0.2 & 1.0 & 1.5 \\ \hline
        CIFAR10 & 0.852$_{\pm{0.02}}$ & 0.851$_{\pm{0.03}}$ & 0.850$_{\pm{0.02}}$ & 0.851$_{\pm{0.03}}$\\
        Tiny-ImageNet & 0.713$_{\pm{0.02}}$ & 0.729$_{\pm{0.01}}$ & 0.731$_{\pm{0.01}}$ & 0.726$_{\pm{0.01}}$\\
        \hlineB{2.5}
        \end{tabular}
    }
	\vspace{0.1cm}
	\caption{AUROC score with varying ratios of $\mathcal{L}_{reg}$.}
	\label{table_regratio}
\end{table}
\endgroup
\vspace{-0.2cm}
\section{Implementation details}
DIAS is an end-to-end framework that all components are learned from the scratch.
For the Copycat and the classifier, we use vanilla CNN~\cite{neal2018osrci}, which is composed of 9 convolution layers.
For the subgroups of convolutional layers, each group contains three 3x3 convolution layers.
Additionally, the backbone network for the generator and the discriminator each contains 4 convolutional layers. 
Moreover, we adopt multi-batch normalization layers to process generated images from GAN separately, as we hope to prevent the problem from distribution mismatch, following \cite{chen2021arpl}.
Note that features from the Copycat do not need to be processed separately.
For scaling parameters, we fix both $\lambda$, and $\beta$ to 0.1.


\end{document}